\documentclass[10pt,twocolumn,letterpaper]{article}
\usepackage{iccv}
\usepackage{graphicx}
\usepackage{ulem, multirow}
\usepackage{caption, CJKutf8}

\definecolor{iccvblue}{rgb}{0.21,0.49,0.74}
\usepackage[pagebackref,breaklinks,colorlinks,allcolors=iccvblue]{hyperref}

\title{TransiT: Transient Transformer for Non-line-of-sight Videography}

\author{%
     \bf Ruiqian Li\textsuperscript{1}\thanks{Equal Contribution}\hspace{2pt}
  ~~ \bf Siyuan Shen\textsuperscript{1,2}$^\ast$
  ~~ \bf Suan Xia\textsuperscript{1}$^\ast$
  ~~ \bf Ziheng Wang\textsuperscript{1}
  ~~ \bf Xingyue Peng\textsuperscript{1}\\
\bf Chengxuan Song\textsuperscript{1}
  ~~ \bf Yingsheng Zhu\textsuperscript{1}
  ~~ \bf Tao Wu\textsuperscript{3}
  ~~ \bf Shiying Li\textsuperscript{1,4}
  ~~ \bf Jingyi Yu\textsuperscript{1,4}\thanks{Corresponding Authors}\vspace{0.3cm}
  \\
  \textsuperscript{1}School of Information Science and Technology, ShanghaiTech University\\ ~~~
  \textsuperscript{2}Lingang Laboratory, Shanghai\\ ~~~
  \textsuperscript{3}School of Physical Science and Technology, ShanghaiTech University\\ ~~~
  \textsuperscript{4}Shanghai Engineering Research Center of Intelligent Vision and Imaging\\ ~~~
  \vspace{-25pt}
}

\begin{document}
\maketitle
\newcommand{\shensy}[1]{{\color{magenta}{[Siyuan: #1]}}}

\newcommand{\wzh}[1]{{\color{teal}{[Ziheng: #1]}}}
\newcommand{\lrq}[1]{{\color{red}{[Ruiqian: #1]}}}
\newcommand{\Li}[1]{{\color{blue}{[Li: #1]}}}

\newcommand{\transients}{\tau}
\newcommand{\ill}{\mathbf{o}}
\newcommand{\wall}[2][2]{\bar{\mathbf{x}}_{#1}^{#2}}
\newcommand{\dis}[2][2]{\mathrm{d}_{#1}^{#2}}
\newcommand{\timee}[2][2]{t_{#1}^{#2}}

\begin{abstract}
High quality and high speed videography using Non-Line-of-Sight (NLOS) imaging benefit autonomous navigation, collision prevention, and post-disaster search and rescue tasks. Current solutions have to balance between the frame rate and image quality. High frame rates, for example, can be achieved by reducing either per-point scanning time or scanning density, but at the cost of lowering the information density at individual frames. Fast scanning process further reduces the signal-to-noise ratio and different scanning systems exhibit different distortion characteristics. In this work, we design and employ a new Transient Transformer architecture called TransiT to achieve real-time NLOS recovery under fast scans. TransiT directly compresses the temporal dimension of input transients to extract features, reducing computation costs and meeting high frame rate requirements. It further adopts a feature fusion mechanism as well as employs a spatial-temporal Transformer to help capture features of NLOS transient videos. Moreover, TransiT applies transfer learning to bridge the gap between synthetic and real-measured data. In real experiments, TransiT manages to reconstruct from sparse transients of $16 \times 16$ measured at an exposure time of 0.4 ms per point to NLOS videos at a $64 \times 64$ resolution at 10 frames per second. We will make our code and dataset available to the community.
\end{abstract}
  
\section{Introduction}

\begin{figure*}[!t]
\centering
\vspace{-20pt}
\includegraphics[width=1.0\textwidth]{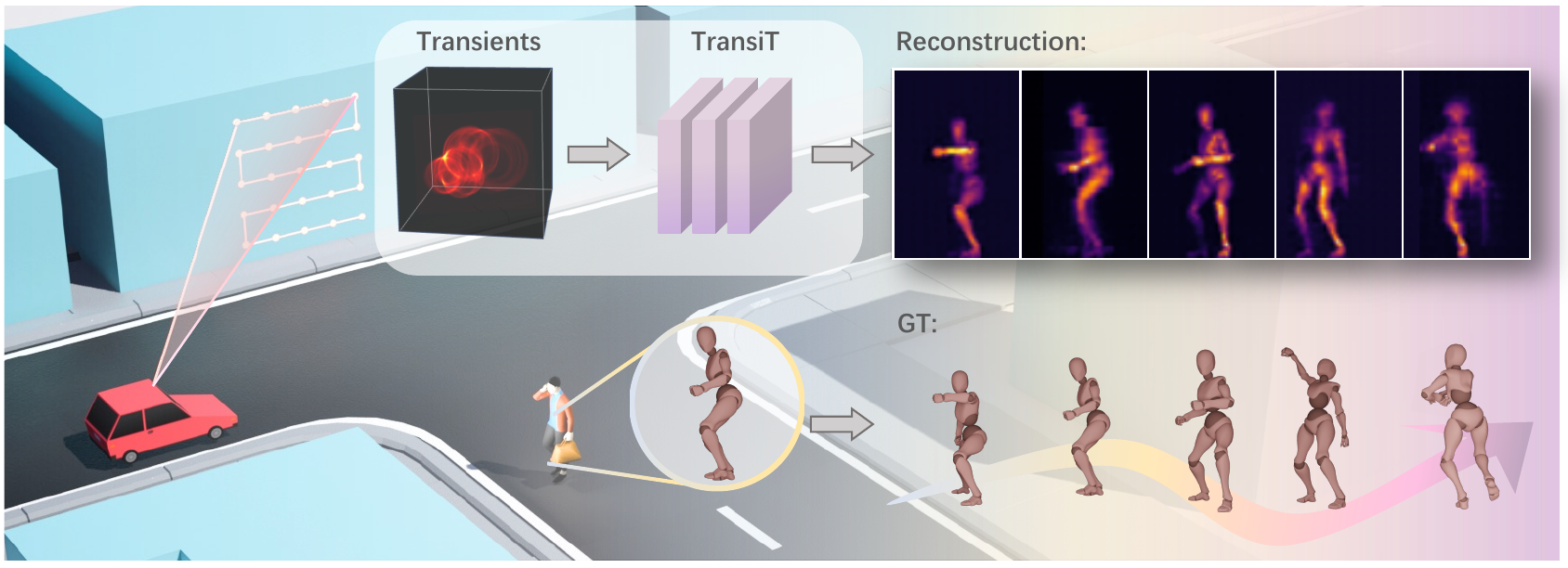}
\vspace{-20pt}
\caption{\textbf{NLOS videography.} An NLOS imaging system captures transients of a moving hidden object (e.g., a walking person) via a relay surface. TransiT is capable of reconstructing high quality NLOS video of a person at 10 FPS using fast and sparse scanning.
\vspace{-10pt}
 }
\label{fig:teaser}
\end{figure*}

Human and machine vision capture dynamic information from the surrounding environment and interpret these variations to avoid hazards and make timely decisions. However, a moving object does not necessarily lie within the direct line-of-sight (LOS), e.g., a pedestrian at corners may be occluded by buildings or other obstacles. The capabilities to detect and recognize such hidden moving targets in scenarios as such are crucial in domain specific applications, ranging from robotics and autonomous driving to post-disaster rescue efforts.

We observe that hidden targets can potentially be captured using a classic non-line-of-sight (NLOS) setup, where an intermediate surface, either naturally existing or intentionally placed, serves as a relay wall~\cite{geng2021recent, faccio2020non}. Fig.~\ref{fig:teaser} illustrates a typical NLOS scenario under a confocal setting, which simplifies the classic setup from a five-dimensional (5D) configuration into 3D~\cite{otoole2018LCT}. After a laser beam illuminates a point on the relay wall, a portion of the photons bounce off and scatter in a spherical wavefront onto a hidden object. A single-photon detector then records a transient, including the number of photons that arrive back at the same point over a specific time interval.  Although this scheme is originally designed to recover static objects, it can be extended to reconstruct dynamic scenes, by applying frame-by-frame methods. Lindell et al.~\cite{lindell2019fk} pioneered the reconstruction of NLOS videos at a resolution of $32 \times 32$ whereas Liu et al.~\cite{liu2020NC} and their follow-up work~\cite{nam2021NC} used a higher resolution scan of $181 \times 131$. These methods require a high sampling rate to ensure high resolution results, which significantly reduces the frame per second (FPS), leading to artifacts such as motion blur and discontinuity. 

SPAD arrays allow multiple-pixel detection in a single shot and can improve the FPS, however, they face many issues such as crosstalk, as well as high computation and memory costs~\cite{pei2021dynamic, young2024enhancing, nam2021NC}. Alternatively, several attempts aim to reduce per point scanning time by using a single-pixel SPAD-based setup ~\cite{buttafava2015non, metzler2021keyhole, pan2022onsite}. Lindell et al.~\cite{lindell2019fk} raster scan a uniform grid of $64 \times 64$ at 2 FPS on a full relay wall of 2 m $\times$ 2 m. Isogawa et al.~\cite{isogawa2020efficient} enhanced the FPS by adopting a circular scanning strategy within a small scanning range. Ye et al.~\cite{ye2024plugandplay} exploit transients of $16 \times 16$ and recover an NLOS video at 4 FPS by reducing the sampling density and incorporating plug-and-play (PnP) regularization and compressed sensing priors. To improve spatial resolution, recent works resort to learning-based approaches~\cite{wang2023non,li2023deep} to upsample transients from $8 \times 8$ and $16 \times 16$ to $32 \times 32$. They are designed primarily for static NLOS reconstruction, and shows limited performance in dynamic scenarios.

We observe that videos captured by an NLOS system in the end are still videos. Recent video transformers have demonstrated strong capabilities in understanding video contents and conducting feature extractions ~\cite{liu2022video,arnab2021vivit, bertasius2021space, wang2022spatial}. Yet, unlike traditional videos, NLOS videos capture the transients of light that do not readily represent meaningful contents. In this paper, we aim to achieve NLOS videography by reconstructing dynamic frames with an initial scanning resolution of 16 $\times$ 16. We propose a Transient Transformer technique called TransiT to achieve real-time NLOS recovery under fast scans. Prior methods upsample sparse transients into virtual dense measurements for high resolution reconstruction~\cite{wang2023non,li2023deep}. In contrast, TransiT directly compresses the temporal dimension of input transients to extract features, reducing computation costs and meeting high FPS requirements. However, this scheme introduces an issue for recovering fine details from sparse spatial-temporal signals. Therefore, TransiT designs a feature fusion mechanism to combine each frame with the features learned from the previous frame and then employs spatial-temporal attention via a Transformer to help capture the spatial-temporal features of NLOS transient videos. 

To deploy our solution to real systems, we observe NLOS transients measured from a SPAD-based NLOS imaging system are also influenced by the image processing pipeline and hardware specifications, e.g., the laser, the galvanometer, and other devices. As a result, the measurements are often convolved with complex and heavy distortions, in particular in per point fast scanning. We thus formulate a new distortion model and construct a large-scale synthetic dataset of dynamic NLOS scenes. We select a variety of over 2K motion sequences from a public dataset~\cite{mixamo} as well as include different motion styles, such as translation, rotation, body movements, etc. Our synthetic dataset contains nearly 200k frames of dynamic NLOS scenes. To bridge the gap between real and synthetic data, it is essential to align the features from both datasets. We design a maximum mean discrepancy (MMD)-based transfer learning method to fine-tune TransiT on real measurements, further enhancing its performance. To validate our technique, we conduct extensive experiments on both synthetic and real-measured datasets. The results demonstrate that TransiT enables us to achieve high quality NLOS videography, converting fast-scan frames at a raw spatial resolution of 16 $\times$ 16 to 64 $\times$ 64 at 10 FPS.
\section{Related Work}
\label{sec:related}

Pioneered by the seminar works \cite{kirmani2009, gupta2012reconstruction, velten2012NC}, time-resolved NLOS imaging has achieved remarkable progress. Back-projection (BP) based algorithms, such as filtered BP \cite{velten2012NC,buttafava2015non} and fast BP \cite{arellano2017fast}, project each transient onto voxels of an NLOS scene and solve the inverse problem based on the ellipsoidal forward model.
O'toole et al.~\cite{otoole2018LCT} present the confocal setting to simplify the forward model from 5D into 3D and the light-cone transform (LCT) algorithm to solve the inverse problem as a 3D deconvolution process for fast reconstruction. The confocal setting and LCT become standard solution for NLOS imaging and benefit recent NLOS research~\cite{xin2019theory, young2020dlct, isogawa2020efficient, shen2021NeTF, li2023deep, mu2022NLOS3d}.
Because the signals detected are weak and are convolved into heavy noise of NLOS imaging system and process, the majority of approaches, including optimization-based~\cite{tsai2019beyond, heide2019non}, wave-based~\cite{lindell2019fk, liu2019PF, liu2020NC}, Fermat flow~\cite{xin2019theory}, and learning-based~\cite{chopite2020deep, chen2020learned, shen2021NeTF, yu2023enhancing}, require dense transient measurements as input to reconstruct high resolution images of the NLOS scene. By treating each frame of a moving hidden object as static, these approaches are capable of recovering a complete NLOS video frame by frame~\cite{lindell2019fk, nam2021NC}. This video may be reconstructed in a low frame rate and without related information between adjacent frames, resulting in artifacts, such as motion blur and discontinuity. From either dense or sparse measurements, our framework enables us to learn related information of moving objects from neighboring frames and reconstruct spatial details of a single frame and alleviate motion blur in NLOS videos.

For dynamic NLOS imaging, many attempts aim to accelerate transient measurement. Streak cameras can scan a line of area on the relay wall by a single shot with short exposure time~\cite{feng2021toward,feng2021ultrafast}, but they are expensive for most practical applications. Owing to good balance in sensitivity and cost, SPAD cameras are the tool of choice to simultaneously record transients of multiple pixels~\cite{pei2021dynamic,nam2021NC,Zhang2024}. However, SPAD cameras are currently under development and only support non-confocal configuration and and limited pixels. Alternatively, single SPAD-based NLOS imaging systems can raster scan the relay wall in arbitrary sampling numbers and patterns~\cite{pan2022onsite, lindell2019fk, isogawa2020efficient}. For fast measurement, many works reduce scanning density~\cite{ye2024plugandplay}, per-point exposure time~\cite{lindell2019fk}, or scanning area~\cite{isogawa2020efficient}. Recently, Ye et al.~\cite{Ye2024} reach 4 FPS for complex dynamic real-life scenes. Through a keyhole, Metzler et al.~\cite{metzler2021keyhole} acquire transients of a moving object from a single light path and estimate the trajectory of the object. We exploit a tailored scanning strategy to reduce per point capture time and scanning density and achieve approximately 102 ms per frame of $16 \times 16$.
 
From sparse transients measured in per point scanning, recent optimization-based~\cite{ye2021compressed,liu2023fewshot,liu2023NC} and learning-based~\cite{wang2023non,li2023deep} approaches attempt to reconstruct high resolution images of hidden objects. Using a superresolution network, Wang et al.~\cite{wang2023non} recover dense virtual measurements from sparse input and exploit existing algorithms for fast reconstruction. Their technique is potentially able to recover an NLOS video frame by frame. Ye et al.~\cite{ye2024plugandplay} reconstruct NLOS videos of moving hidden objects and filter image and motion blurring effects as noise. In contrast to these methods, we train TransiT on a large-scale synthetic dataset, which composes a variety of dynamic NLOS scenes, and leverage relevant information of diverse moving hidden objects for high quality NLOS videography.
\section{Distortion Model Under Fast Scanning}\label{sec:fsd}

To achieve NLOS videography at 10 FPS, we use a confocal imaging system~\cite{otoole2018LCT} and scan a $16\times 16$ grid on the relay wall employing a serpentine scanning pattern, with approximately 0.4 ms scanning time per point. 
 
\begin{figure}[t]
\centering
\includegraphics[width=0.48\textwidth]{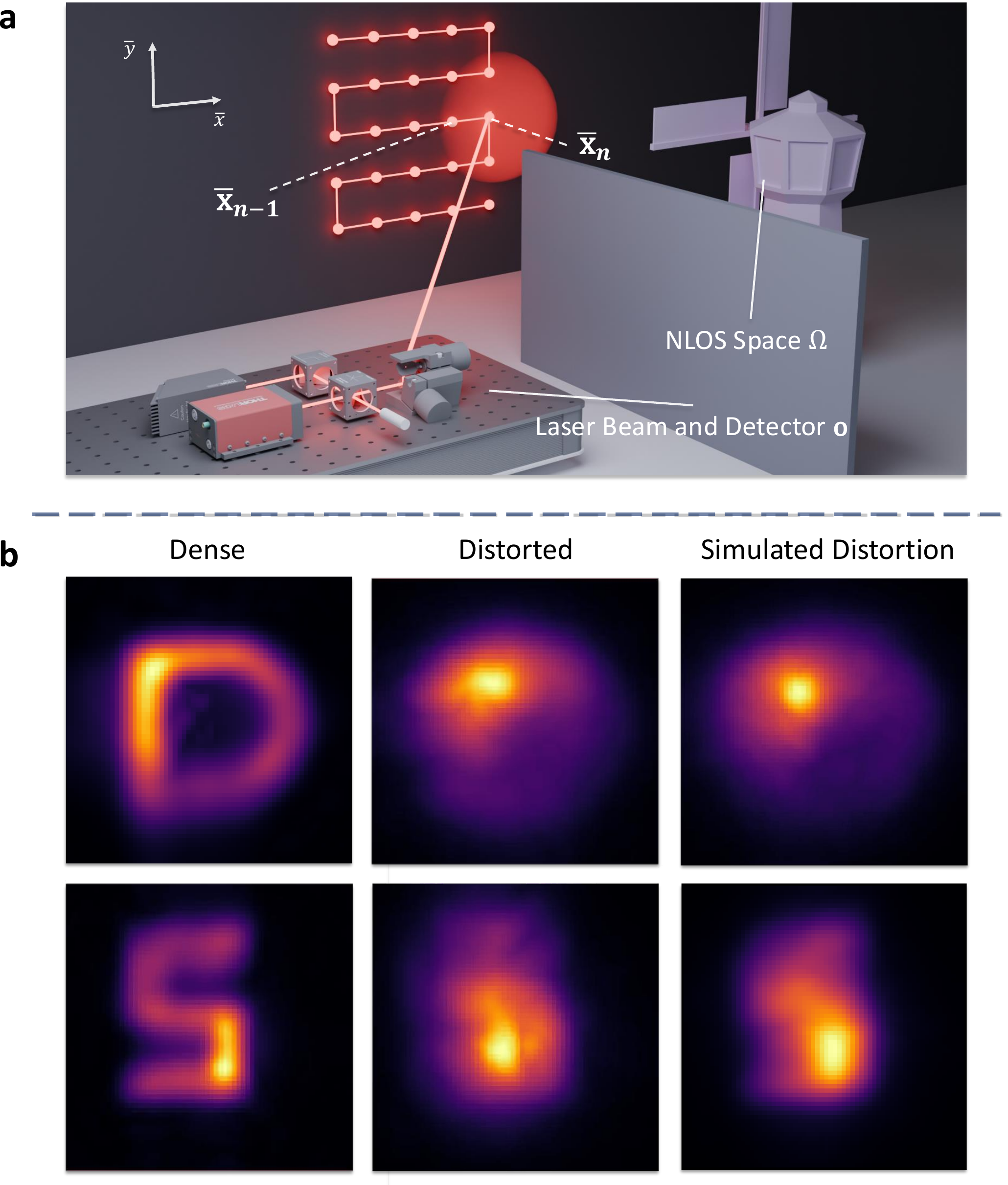} 
\vspace{-20pt}
\caption{\textbf{Distortion model under fast scanning.} 
(a) System configuration for distortions under fast scanning. Due to the limited galvanometer's speed, illumination and detection scan on the relay wall in a linear form rather than at a single point. (b) Images reconstructed using f-k~\cite{lindell2019fk} from three different transients as input. Dense: $64\times 64$ grid with 2 ms per point, distortion-free. Distorted: $16\times 16$ grid with 0.4 ms per point. Simulated Distortion: $16 \times 16$ grid generated by applying our distortion model to the transients of scanning points picked from the Dense.}
\label{fig:imaging}
\vspace{-10pt}

\end{figure}

Following the forward model~\cite{otoole2018LCT}, the ideal transients $\tau(\bar{\mathbf{x}}_n, t)$ in Fig.~\ref{fig:imaging}(a) can be formulated as:
{\vspace{-5pt}
\small
\begin{equation}
    \tau(\bar{\mathbf{x}}_n, t) = \frac{1}{r^4} \iiint_{\Omega} \rho(\mathbf{x})\cdot \delta(2\|\bar{\mathbf{x}}_n - \mathbf{x}\| - tc)\,d\mathbf{x},
    \label{eq:simp}
    \vspace{-5pt}
\end{equation}
}
\noindent where $\Omega$ represents the NLOS space and $\rho$ the albedo of the hidden scene at any point $\mathbf{x} = (x, y, z)$. The Dirac delta function $\delta$ converts the distance to time $t=2\|\bar{\mathbf{x}}_n-\mathbf{x}\| / c$ from the illumination point to the hidden scene and back to the detection point, with $c$ the speed of light.

Ideally, we assume that the laser beam moves instantly from point to point inside the scanning grid and can neglect the time of its movement between scanning points. However, the direction of the laser beam is guided by a scanning galvanometer system, which has a minimum response time of $\sim $0.4 ms from when it receives a signal to when it completes the movement. In contrast to the ideal per-point scanning,  this response time creates a continuous scanning path. Ideal transients $\hat{\tau}(\bar{\mathbf{x}}_n, t)$ recorded under fast scanning are therefore an integral of photon events illuminated along the path between points. Denote $\wall[n]{m}$ as the point on the wall located at $\bar{\mathbf{x}}_n$ being shifted a distance $m$ towards the previous scanning point, we have:
{\vspace{-5pt}
\small
\begin{equation}
    \hat{\tau}(\bar{\mathbf{x}}_n, t) = 
    \frac{1}{\|S\|}\int_{S}\tau\left(\wall[n]{\mathbf{s}}, t\right)\,d\mathbf{s}, 
    \label{eq:distort_simp}
    \vspace{-5pt}
\end{equation}
}
\noindent where $S=\bar{\mathbf{x}}_n - \bar{\mathbf{x}}_{n-1}$ represents the one-dimensional path between adjacent scanning points, $\frac{1}{\|S\|}$ the normalization term. We assume $\|S\|\neq 0$, otherwise $\hat{\tau}(\bar{\mathbf{x}}_n, t)=\tau(\bar{\mathbf{x}}_n, t)$. The serpentine scanning pattern restricts adjacent points to change along either $\bar{x}$- or $\bar{y}$-axis on the relay wall, allowing us to simplify this path integral to a single dimension.

\begin{figure*}[h]
\centering
\vspace{-20pt}
\includegraphics[width=1.0\textwidth]{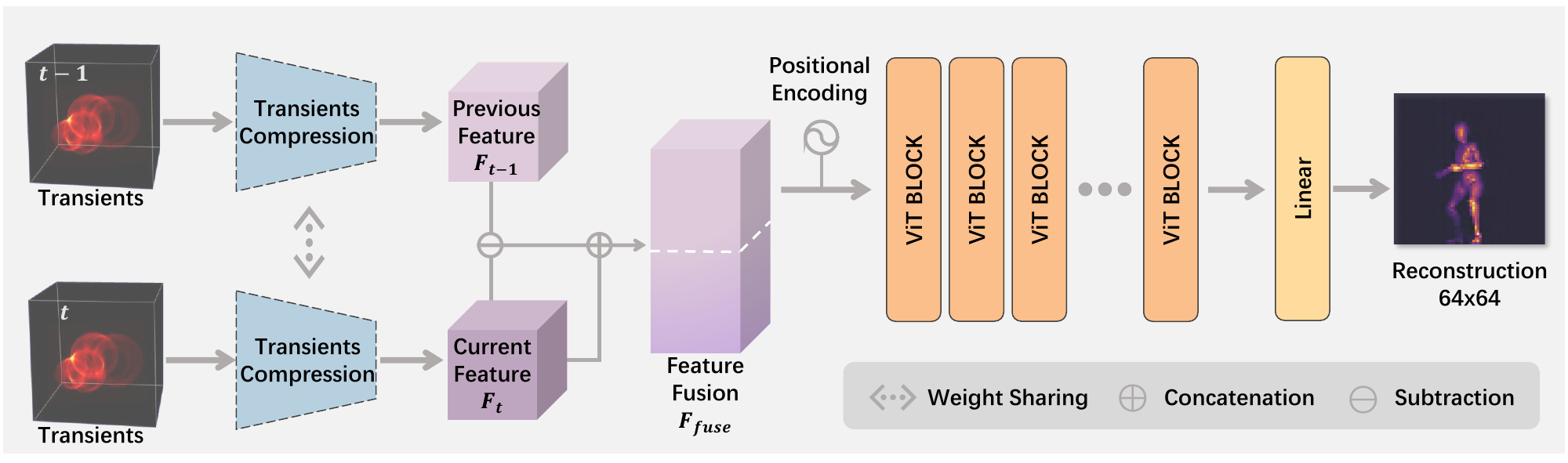} 
\vspace{-15pt}

\caption{\textbf{Pipeline of TransiT.} 
TransiT is a transformer-based architecture with the transients of current and previous frames as input. Transient compression extracts features by compressing the input transients along the temporal axis. Feature fusion combines the current frame's features with the difference between the current and previous frame features. ViT blocks with spatial-temporal attention then process the fused features. Followed by a linear layer, TransiT outputs a high resolution reconstruction.}
\vspace{-10pt}
\label{fig:pipline}
\end{figure*} 

Real measurement records two additional photon propagation paths compared to the ideal transients in Eq.~\ref{eq:simp}: One from the laser to the relay wall and one from the relay wall to the detector. Treating the laser and detector as being at the same location $\mathbf{o} = (x_o, y_o, z_o)$, defining $\dis[n]{m} = \|\wall[n]{m} - \mathbf{o}\|$ and $\dis[n]{} = \dis[n]{\mathbf{0}}$, the real measurement is:
{\vspace{-5pt}
\small
\begin{equation}
    \label{eq:simp2orig}
    \tau_*(\bar{\mathbf{x}}_n, t) = \frac{1}{{\dis[n]{}}^2}\cdot \tau\left(\bar{\mathbf{x}}_n, t-\frac{2{\dis[n]{}}}{c}\right). 
    \vspace{-5pt}
\end{equation}
}
\noindent where $1/{\dis[n]{}}^2$ is the attenuation from diffuse reflection after photons arrive at the detection point. Combining Eq.~\ref{eq:distort_simp} and Eq.~\ref{eq:simp2orig}, the real measurement under fast-scan scenario $\hat{\tau}_*(\bar{\mathbf{x}}_n, t)$ can be derived as:
{\vspace{-5pt}
\small
\begin{equation}
    \hat{\tau}_*(\bar{\mathbf{x}}_n, t) = \frac{1}{\|S\|}\int_{S} \left(\frac{\dis[n]{\mathbf{s}}}{\dis[n]{}}\right)^2\cdot \tau_*\left(\wall[n]{\mathbf{s}}, t + \frac{2(\dis[n]{\mathbf{s}} - \dis[n]{})}{c}\right)\,d\mathbf{s},
    \label{eq:distort_meas}
\end{equation}
}   
\noindent where $\wall[n]{\mathbf{s}}$ represents a specific point on the path between $\wall[n]{}$ and $\wall[n-1]{}$, while $\dis[n]{\mathbf{s}}$ is the directed line between $\wall[n]{\mathbf{s}}$ and the detector. 
We need to convert the real measurement to a scanning grid, Eq.~\ref{eq:distort_meas} thus reveals the distortion introduced by the mismatching between the focal point and the scanning grid.
Rewriting Eq.~\ref{eq:distort_meas} using Eq.~\ref{eq:simp2orig}, the distortion in ideal transients with real measurement error becomes:
{\vspace{-5pt}
\small
\begin{equation}
    \hat{\tau}(\bar{\mathbf{x}}_n, t) = \frac{1}{\|S\|}\int_{S} \left(\frac{\dis[n]{}}{\dis[n]{\mathbf{s}}}\right)^2\cdot \tau\left(\wall[n]{\mathbf{s}}, t + \frac{2(\dis[n]{} - \dis[n]{\mathbf{s}})}{c}\right)\,d\mathbf{s}.
    \label{eq:final}
    \vspace{-5pt}
\end{equation}
}   
This integral can then be discretized to:
{\vspace{-5pt}
\small
\begin{equation}
\label{eq:discrete}
\hat{\tau}(\wall[n]{}, t) \approx \frac{1}{\|S\|}\sum_{i=1}^{M} \left(\frac{\dis[n]{}}{\dis[n]{i\Delta \mathbf{s}}}\right)^2\cdot \tau\left(\wall[n]{i\Delta \mathbf{s}}, t + \frac{2(\dis[n]{} - \dis[n]{i\Delta\mathbf{s}})}{c}\right) \Delta \mathbf{s}, 
\end{equation}
}
\noindent where $M$ represents the number of sampled points between two adjacent scanning points, and $\Delta s$ the distance between two sample points (please see Supplementary Materials for all detailed derivations above). Fig.~\ref{fig:imaging}  (b) shows the images reconstructed from distortion-free, real-world distorted, and simulated distorted data. 

Eqs.~\ref{eq:final} and ~\ref{eq:discrete} show the distortion model mapping from per-point scanned dense ideal transients to fast-scan sparse ideal transients. However, the inverse problem in this context is difficult to solve mathematically, i.e., we cannot recover undistorted transients from distorted. Therefore, when training TransiT, we generate synthetic data with high spatial resolution (e.g., $64\times 64$) and intentionally introduce fast-scan distortions by applying Eq.~\ref{eq:final}, allowing TransiT to reconstruct the hidden scene from distorted data.
\section{TransiT Architecture} 
\label{sec:method}

Our TransiT tackles the challenges of high frame rate NLOS video reconstruction of dynamic scenes. 
The core innovation of TransiT lies in its ability to utilize the generative capabilities inherent in transformer, enabling the recovery of NLOS video from low-resolution data with fast-scan distortion. Fig.~\ref{fig:pipline} shows the overall pipeline of TransiT.

\textbf{Transients compression.}
A crucial challenge in high frame rate NLOS Video reconstruction is balancing the need for high temporal resolution while preserving enough spatial detail for effective reconstruction. Unlike previous methods that rely on upsampling techniques to increase the resolution of input transients, our approach directly compresses the input transients using a linear layer to extract transients feature $\mathcal{F}$. Specifically, we compress the temporal axis of the input transients (represented as histograms) into a lower-dimensional feature space, significantly reducing the input complexity. Despite the original NLOS histograms spanning hundreds or even thousands of time bins, experiments demonstrate that compressing them into a 32-dimensional feature vector does not degrade performance. Intuitively, in dynamic and rapidly scanned NLOS scenes, the most valuable information from each histogram is often concentrated in one or a few peak positions and amplitudes.

\textbf{Feature fusion.}
Another challenge in NLOS video reconstruction arises from the difficulty in capturing fine details of dynamic scenes from sparse data. To address this, we introduce a feature fusion technique that emphasizes temporal differences between consecutive frames, highlighting critical dynamic changes in the scene. Specifically, we compute the difference between the features of the current frame transients $\mathcal{F}_{t}$ and the previous frame $\mathcal{F}_{t-1}$ and concatenate it with the current frame’s features:
\vspace{-5pt}
\begin{equation}
    \mathcal{F}_{fuse} = concat(\mathcal{F}_{t}, \mathcal{F}_{t} - \mathcal{F}_{t-1}).
    \vspace{-5pt}
\end{equation}
This fused feature $\mathcal{F}_{fuse}$ combines both the current state of the scene and the observed temporal changes, providing a richer representation for reconstruction.

\textbf{Transformer with spatiotemporal attention.}
To model the complex spatiotemporal dependencies in NLOS video reconstruction, we utilize Vision Transformer (ViT) blocks with spatial and temporal positional encodings. 
The spatial positional encoding $PE^{spatial}$ is two-dimensional, representing the $x$ and $y$ coordinates of the sampling points. Temporal positional encoding $PE^{time}$, on the other hand, is one-dimensional  the distinct frames of the transients sequence. Spatial and temporal positional encodings are combined via element-wise summation and subsequently applied to the fused feature $\mathcal{F}_{fuse}$. The resulting encoded feature is then passed through the transformer blocks $\mathcal{B}$ with a simple linear to generate the final image $I$ of current frame:
\vspace{-5pt}
\begin{equation}
I = \mathcal{B}(\mathcal{F}_{fuse} + PE^{spatial} + PE^{time}).
\vspace{-5pt}
\end{equation}

\textbf{Training.}
We use a two-stage training approach to optimize the model. In the first stage, we minimize the mean squared error (MSE) between the ground truth image \( I_{GT} \) and the predicted images from the network:
\vspace{-5pt}
\begin{equation}
\mathcal{L}_{imaging} = \left\| I_{GT}- I \right\|^2 .
\label{eq:loss1}
\vspace{-5pt}
\end{equation}

A key challenge in NLOS video reconstruction is the distribution gap between synthetic and real-measured data, as system responses and noise levels vary across hardware platforms, and external factors can distort the data. To address this, we propose a second training stage that employs Maximum Mean Discrepancy (MMD) Loss which is used in transfer learning to measure the discrepancy between the distributions of synthetic and real-measured data. Specifically, we extract fused features from both synthetic and real-measured data to calculate the MMD loss between them:
\vspace{-5pt}
\begin{equation}
    \mathcal{L}_{MMD} = \left\| \frac{1}{n} \sum_{i=1}^{n} \phi(\mathcal{F}^i_{real}) - \frac{1}{m} \sum_{j=1}^{m} \phi(\mathcal{F}^j_{syn}) \right\|^2,
    \vspace{-5pt}
\end{equation}
where where $n$ and $m$ represent the number of feature samples extracted from the real-measured and synthetic datasets, respectively. \( \phi(\cdot) \) is the feature map corresponding to a Gaussian kernel. The total training loss for stage two combines MMD loss with image reconstruction loss:
\vspace{-5pt}
\begin{equation}
\mathcal{L}_{total} = \mathcal{L}_{imaging} + \lambda \mathcal{L}_{MMD},
\label{eq:loss2}
\vspace{-5pt}
\end{equation}
where \( \lambda \) is a hyperparameter that controls the trade-off between reconstruction accuracy and domain alignment. 

While this training strategy provides an effective solution to mitigate domain gaps, it remains an optional strategy rather than a mandatory requirement. In practical NLOS scenarios, discrepancies between synthetic and real-world settings may arise due to material properties, hardware characteristics, and noise artifacts. For instance, most simulations assume that both the relay surface and the hidden object exhibit ideal diffuse reflectance, whereas real-world conditions may introduce specular reflections. Additionally, factors such as laser source variations and SPAD-specific noise introduce distortions that are sometimes difficult to model precisely in simulation.

A key advantage of MMD Loss in such cases is that it's self-supervised, requiring only real-world measurement without the need for corresponding ground truth, which is usually difficult to obtain. Moreover, MMD-based domain adaptation can achieve good performance with significantly fewer real-world samples than synthetic data, further enhancing its practical applicability. In cases where synthetic and real-world conditions are already well-aligned, this two-stage training strategy may be unnecessary. Further training details are provided in the next section.
\section{Experiments}
\label{sec:exp}
\subsection{Experimental Setup}

\begin{figure}[t]
\centering
\includegraphics[width=0.5\textwidth]{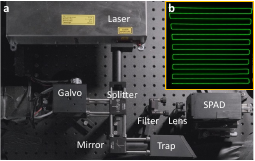} 
\vspace{-20pt}
\caption{\textbf{System setup.} (a) Our NLOS imaging system, and (b) The fast scanning pattern.}
\vspace{-15pt}

\label{fig:system}
\end{figure}

\begin{figure*}[!t]
\centering
\vspace{-20pt}
\includegraphics[width=1.0\textwidth]{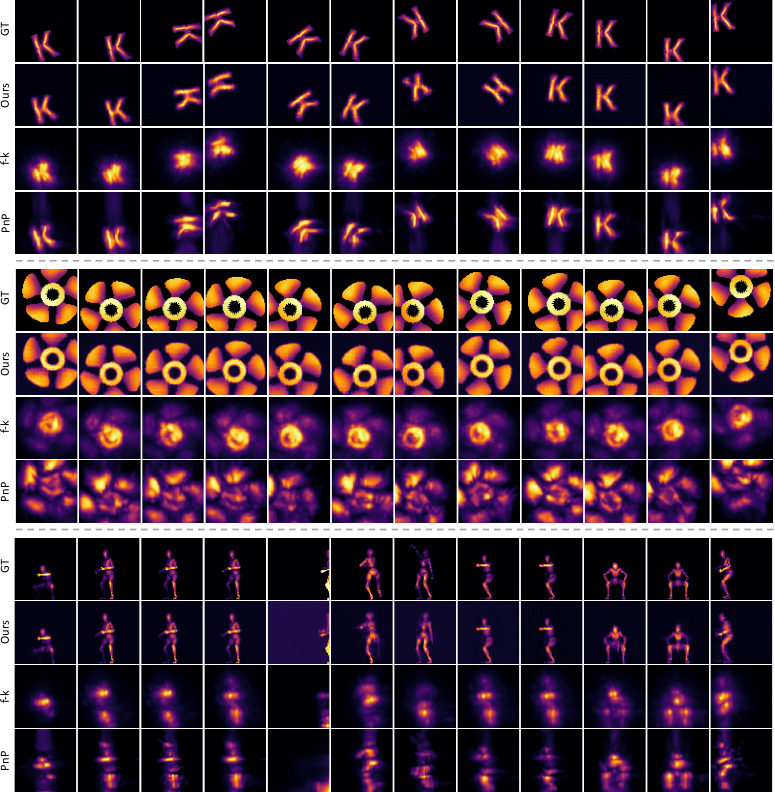} 
\vspace{-15pt}
\caption{\textbf{Comparison of synthetic results.} From top to bottom: Ground truth, ours, f-k and PnP. The results are reconstructed across multiple frames of 16×16 of noisy synthetic data for different objects — a character 'K', a propeller, and a human.}
\vspace{-10pt}

\label{fig:simulation}
\end{figure*}

\textbf{Synthetic dataset.}
Based on our distortion model, we construct a large scale synthetic dataset, which includes 10,000 motion sequences and 100,000 frames rendered from a diversity of objects, e.g., letters, windmills, propellers, and full-body and half-body of humans. The object sizes range from 1 m $\times$ 1 m to nearly 2 m $\times$ 2 m. The motion styles vary from simple translation and rotation to complex movements. Each motion sequence spans a duration of 5 seconds, with 10 FPS, yielding 50 consecutive frames. The motion speed is approximately 40 cm/s across motions. We exploit the motion sequences of full-body and half-body of humans from  Mixamo~\cite{mixamo}, a human motion dataset with over 2,000 motion sequences.
We generate the synthetic transients at spatial resolution of 64 $\times$ 64 and at temporal resolution of 20 ps by scanning on a 2 m $\times$ 2 m relay wall. 

\textbf{Real-measured data.} We construct an NLOS imaging system, as illustrated in Fig. \ref{fig:system}, under a confocal configuration. Light from the laser (Katana 05-HP, 532 nm, repetition rate 1 MHz) is directed by a 2D galvanometer (GVS012) and scan on the relay wall. A fast-gated SPAD with a 50 mm lens is coupled with a PicoHarp 300 and a delayer (PSD-MOD) to record transients at time-bin resolution 4 ps. The hardware is positioned 2 m away from the relay wall. 
The system is calibrated with a time jitter of 72 ps using the onsite method \cite{pan2022onsite}. Due to the mechanical speed limitations of the galvanometer, we adopt a serpentine scanning pattern (Fig. \ref{fig:system}(b)) to minimize rapid large-angle changes over short periods. We scan a $16\times 16$ grid covering an area of 1 m $\times$ 1 m, with each point being scanned for 0.4 ms.

\begin{table}[t]
\centering
\small
\caption{Comparison results. We compare the reconstruction performance of different moving objects using different algorithms (f-k, PnP), evaluated by Euclidean Distance (ED), Cosine Similarity (CS), Structural Similarity Index (SSIM), and Peak Signal-to-Noise Ratio (PSNR).}
\vspace{-5pt}
\label{tab:Synthetic}
\begin{tabular}{cccccc}
\hline
\textbf{Object} & \textbf{Method} & \textbf{ED$\downarrow$} & \textbf{CS$\uparrow$} & \textbf{SSIM$\uparrow$} & \textbf{PSNR$\uparrow$} \\
\hline
\multirow{3}{*}{Character} & f-k & 0.1286 & 0.7876 & 0.6677 & 17.86 \\
                               & PnP & 0.0923 & 0.8575 & 0.6764 & 20.77 \\
                               & Ours & \textbf{0.0520} & \textbf{0.9418} & \textbf{0.9227} & \textbf{25.87} \\
\hline
\multirow{3}{*}{Propeller} & f-k & 0.3180 & 0.6854 & 0.1902 & 10.01 \\
                               & PnP & 0.2707 & 0.7800 & 0.2818 & 11.39 \\
                               & Ours & \textbf{0.0904} & \textbf{0.9781} & \textbf{0.8211} & \textbf{20.91} \\

\hline
\multirow{3}{*}{Human} & f-k  & 0.1136 & 0.6265 & 0.5791 & 19.00\\
                               & PnP & 0.1018 & 0.6939 & 0.6706 & 19.92 \\
                               & Ours & \textbf{0.0415} & \textbf{0.9272}  & \textbf{0.8041} & \textbf{29.45} \\
\hline
\end{tabular}
\vspace{-10pt}
\end{table}

\begin{figure*}[t]
\centering

\vspace{-20pt}

\includegraphics[width=1.0\textwidth]{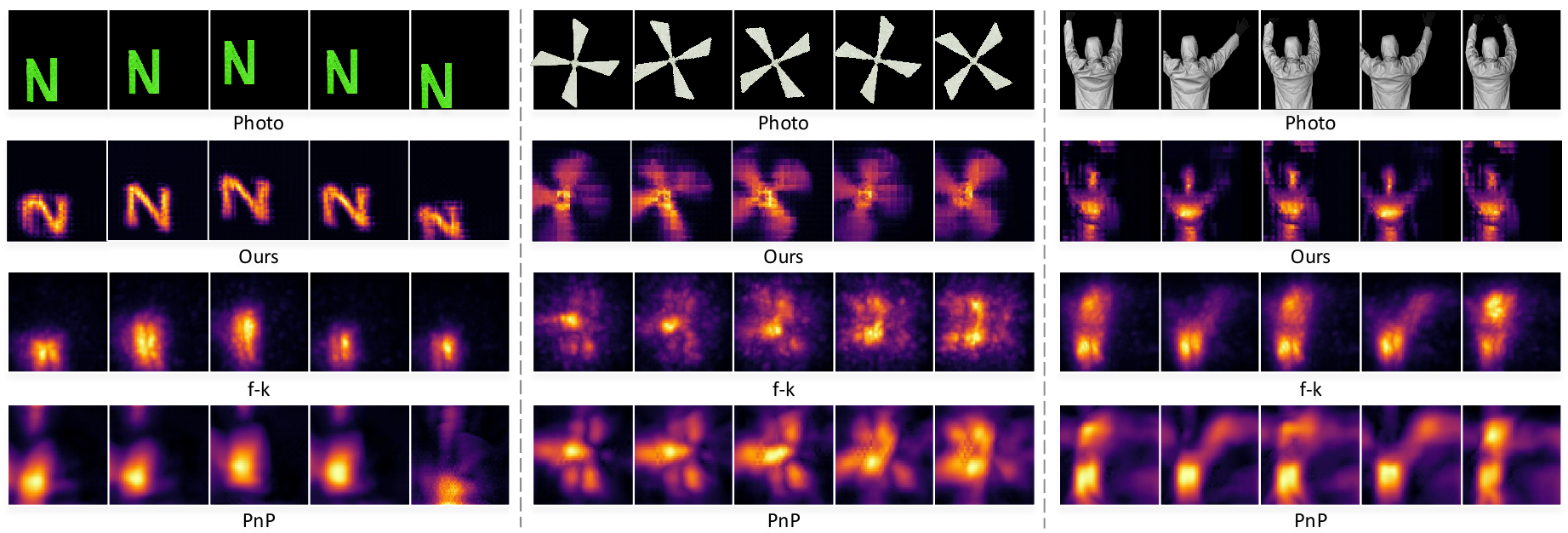}
\vspace{-15pt}
\caption{\textbf{Comparison of real-measured results.} From top to bottom: Ground truth, ours, f-k, and PnP. The results are reconstructed across multiple frames of $16\times 16$ of real-measured data for different objects — a character 'N', a windmill, and a human.}

\vspace{-10pt}
\label{fig:measurement}
\end{figure*}

\textbf{Implementation.}
We implement TransiT using PyTorch, employing the AdamW optimizer with hyperparameters $\beta_1=0.9$, $\beta_2=0.95$, and weight decay set to $0.01$. The learning rate follows a cosine decay schedule, starting from $5 \times 10^{-3}$ to $1 \times 10^{-4}$, with a linear warmup over the first 10 epochs.
The input transients, originally of size $16 \times 16 \times T$, are first compressed to a 32-dimensional latent representation on $T$. The reconstruction network consists of 8 ViT blocks, each with 8 attention heads. We employ FlashAttention~\cite{dao2022flashattention} to improve memory efficiency and speed during training and inference.
For training, we use a batch size of 64 per GPU and distribute training across 24 Nvidia A800 GPUs. The total number of training epochs is set to 1000, and it requires approximately 24 hours to optimize with Eq.~\ref{eq:loss1}. For real experiments, we fine-tune TransiT for an additional 100 epochs using 200 real NLOS transient frames on 8 Nvidia A800 GPUs and it requires approximately 2 hours to optimize with Eq.~\ref{eq:loss2} with $\lambda=0.01, n=32, m=64$. Training is conducted with mixed precision using PyTorch AMP to accelerate computations.
For inference, we deploy the model on a single Nvidia RTX 3090 GPU. The processing time per frame is approximately 0.6 ms, enabling real-time NLOS video reconstruction at 10 FPS.

\textbf{Comparison methods.}
We carry out experiments on the dataset using three approaches: our TransiT, f-k~\cite{lindell2019fk}, and PnP~\cite{ye2024plugandplay}. To ensure a fair comparison, we standardize the input size to $16\times 16$ and the output size to $64\times 64$, and apply a consistent distortion model across all methods. For the synthetic data, we apply our distortion model to convert distortion-free transients at $64\times 64$ into distorted transients at $16\times 16$. For real-measured data, we directly use our captured transients at $16\times 16$.
For f-k, we upsample the $16\times 16$ input to $64\times 64$ via interpolation, producing an output of $64\times 64$. For PnP, we follow the preprocessing steps from their public code, padding the $16\times 16$ spatial resolution to $64\times 64$ before applying the algorithm.

\subsection{Synthetic Results}
Fig.~\ref{fig:simulation} showcases the multi-frame reconstruction results of our synthetic data. From top to bottom, the sequences depict the character 'K', a propeller, and a full-body of human. Specifically, the motion of the character 'K' involves a combination of translation and rotation, while the propeller is randomly sampled from frames along different rotational sequences. The full-body human is similarly sampled from frames across various motion sequences.

Due to fast-scan distortions, f-k produces highly blurred reconstructions and fails to distinguish the object although it can track the dynamic position. PnP performs better for the character and full-body human, where shape and position are more discernible, but shows less satisfactory performance in the propeller. Our method, in contrast, achieves superior reconstructions by accounting for system hardware noise and fast-scan distortion during training, while leveraging the strong learning and representation capabilities of the Transformer. To quantitatively analyze the performance, we compare the results 
 from different methods across various objects in terms of ED, CS, SSIM, and PSNR. As shown in Table \ref{tab:Synthetic}, TransiT 
 outperforms f-k and PnP. Additional results are included in Supplementary Materials.

\subsection{Real-measured Results}

Fig.~\ref{fig:measurement} presents the multi-frame reconstruction results for three different objects: a character 'N', a windmill  and a half-body human. The size of the planar character is 50 cm $\times$ 50 cm, with a movement speed of approximately 20 cm/s. The windmill has a diameter of around 80 cm, with a rotational speed of approximately 15°/s. Due to the distortions, f-k and PnP produce highly blurred shapes of the objects although they can track the dynamic positions. 

The results on both synthetic and real-measured datasets demonstrate that TransiT enables us to robustly handle complex and heavy distortions. We further conduct experiments on the real-measured data from f-k~\cite{lindell2019fk} and show comparison results in Supplementary Materials, as well as additional results of other objects. We also want to note that there are some block-like artifacts in our real results which may be caused by the materials of the hidden objects. Synthetic objects used for training are diffuse, whereas some real objects are retro-reflective. Handling varied materials is still an open question. We consider this as interesting future work and one could help solve this by incorporating larger NLOS datasets with diverse material and more advanced NLOS foundation models.

\subsection{Ablation Study}
\begin{figure}[t]
\centering
\includegraphics[width=0.477\textwidth]{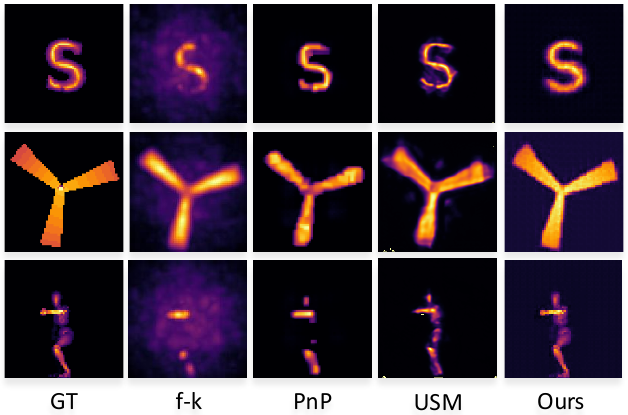} 
\caption{Ablation study. The results are reconstructed from transients of three static objects under sparse scanning.}
\vspace{-10pt}
\label{fig:static}
\end{figure} 
To assess performance of TransiT in the absence of distortions, we carry out an ablation study. Using the pretained TransiT, we fine-tune it on distortion-free data for 100 epochs and test it on transients of three static objects: A character 'S', a propeller, and a human. These transients are captured under sparse scanning. Fig.~\ref{fig:static} shows the comparison results from four methods. f-k yields relatively noisy results with noticeable artifacts whereas PnP can get clear results by incorporating a band-pass filtering and a video denoising network. USM~\cite{li2023deep} generates relatively clear reconstruction results in some cases owing to its transient recovery network, whereas the reconstruction of the letter 'S' still exhibits slight blur. In contrast, TransiT offers superior performance under the scenario without fast-scan distortions.

We further provide quantitative comparisons of four methods in terms of ED, CS, SSIM, and PSNR as shown in Table \ref{tab:ablations}. Both qualitative and quantitative results demonstrate that our method outperforms the others in terms of metrics and handle high quality reconstruction for either static or dynamic NLOS scenes.
\section{Discussion and Conclusion}
\label{sec:conclusion}

We have presented TransiT, a new transformer-based technique, to reconstruct high FPS, high quality non-line-of-sight videos. This new solution allows us to capture sparse transients ($16 \times 16$) at the exposure time 0.4 ms per point using a single-pixle SPAD. Resultant transients involve distortions, which are convolved during the imaging and fast scanning processes. We have formulated a distortion model and presented effective solutions to practically deploy TransiT to real NLOS imaging systems. Comprehensive experiments show that TransiT outperforms the state-of-the-art on both reconstruction quality and frame rates. While our system currently operates at 10 FPS, we could theoretically achieve even higher frame rates without modifying the hardware or network structure. However, we selected 10 FPS to balance reconstruction quality and temporal consistency.

\begin{table}[t]
\centering
\small
\caption{Quantitative results of ablation study. We compare the reconstruction performance of four methods (f-k, PnP, USM) for different static objects in terms of Euclidean Distance (ED), Cosine Similarity (CS), Structural Similarity Index (SSIM), and Peak Signal-to-Noise Ratio (PSNR).} 
\vspace{-5pt}
\label{tab:ablations}  
\begin{tabular}{cccccc}
\hline
\textbf{Object} & \textbf{Method} & \textbf{ED$\downarrow$} & \textbf{CS$\uparrow$} & \textbf{SSIM$\uparrow$} & \textbf{PSNR$\uparrow$} \\
\hline
\multirow{4}{*}{Character} & f-k & 0.1352 & 0.6657 & 0.1224 & 17.37\\
                           & PnP & 0.0763 & 0.8639 & 0.8852 & 22.34 \\
                           & USM & 0.0548 & 0.9317 & 0.9163 & 25.19 \\
                           & Ours & \textbf{0.0531} & \textbf{0.9567} & \textbf{0.9261} & \textbf{25.49} \\
\hline
\multirow{4}{*}{Propeller}  & f-k & 0.1487 & 0.8456 & 0.2365 & 16.55 \\
                           & PnP & 0.1204 & 0.8926 & 0.7817 & 18.38 \\
                           & USM & \textbf{0.1081} & 0.9336 & 0.8016 & \textbf{21.92} \\
                           & Ours & 0.1174 & \textbf{0.9556} & \textbf{0.8308} & 18.61 \\
\hline
\multirow{4}{*}{Human}     & f-k & 0.1525 & 0.4828 & 0.0957 & 16.34 \\
                           & PnP & 0.0754 & 0.6925 & 0.8515 & 22.49 \\
                           & USM & 0.0483 & 0.8641 & 0.9227 & 24.24 \\
                           & Ours & \textbf{0.0241} & \textbf{0.9652} & \textbf{0.9361} & \textbf{26.39} \\
\hline
\end{tabular}
\vspace{-10pt}
\end{table}

While TransiT has made significant progress, there remain a number of challenges in NLOS videography. 
First, fast scanning reduces per point exposure time and the scanning density, resulting in sparse and distorted transients. Existing techniques are difficult to denoise these distortions. A potential solution is to employ continuous scans as inputs for the reconstruction process, rather than sparse and grid-based data. This would require TransiT to be fine-tuned to cope with the new scanning process and data types.   
Second, for complex NLOS scenes, single-pixel SPADs require prolonged acquisition time to capture dense measurements with rich information, which limits the achievable frame rates. Although SPAD arrays are the tool in the future, they are currently constrained by the resolution and the sensitivity. Our TransiT was initially designed to single-point scanning but could be potentially extended to SPAD arrays for achieving even higher frame rates. These are our immediate future work and hopefully a major step towards real-time NLOS video reconstruction in practical applications.

{
    \small
    \bibliographystyle{ieeenat_fullname}
    \bibliography{main}

\begin{thebibliography}{42}
\providecommand{\natexlab}[1]{#1}
\providecommand{\url}[1]{\texttt{#1}}
\expandafter\ifx\csname urlstyle\endcsname\relax
  \providecommand{\doi}[1]{doi: #1}\else
  \providecommand{\doi}{doi: \begingroup \urlstyle{rm}\Url}\fi

\bibitem[mix()]{mixamo}
Mixamo, https://www.mixamo.com/\#/.

\bibitem[Arellano et~al.(2017)Arellano, Gutierrez, and Jarabo]{arellano2017fast}
Victor Arellano, Diego Gutierrez, and Adrian Jarabo.
\newblock Fast back-projection for non-line of sight reconstruction.
\newblock \emph{Open Express}, 25\penalty0 (10):\penalty0 11574--11583, 2017.

\bibitem[Arnab et~al.(2021)Arnab, Dehghani, Heigold, Sun, Lu{\v{c}}i{\'c}, and Schmid]{arnab2021vivit}
Anurag Arnab, Mostafa Dehghani, Georg Heigold, Chen Sun, Mario Lu{\v{c}}i{\'c}, and Cordelia Schmid.
\newblock Vivit: A video vision transformer.
\newblock In \emph{Proceedings of the IEEE/CVF international conference on computer vision}, pages 6836--6846, 2021.

\bibitem[Bertasius et~al.(2021)Bertasius, Wang, and Torresani]{bertasius2021space}
Gedas Bertasius, Heng Wang, and Lorenzo Torresani.
\newblock Is space-time attention all you need for video understanding?
\newblock In \emph{ICML}, page~4, 2021.

\bibitem[Buttafava et~al.(2015)Buttafava, Zeman, Tosi, Eliceiri, and Velten]{buttafava2015non}
Mauro Buttafava, Jessica Zeman, Alberto Tosi, Kevin Eliceiri, and Andreas Velten.
\newblock Non-line-of-sight imaging using a time-gated single photon avalanche diode.
\newblock \emph{Optics express}, 23\penalty0 (16):\penalty0 20997--21011, 2015.

\bibitem[Chen et~al.(2020)Chen, Wei, Kutulakos, Rusinkiewicz, and Heide]{chen2020learned}
Wenzheng Chen, Fangyin Wei, Kiriakos~N Kutulakos, Szymon Rusinkiewicz, and Felix Heide.
\newblock Learned feature embeddings for non-line-of-sight imaging and recognition.
\newblock \emph{ACM Transactions on Graphics}, 39\penalty0 (6):\penalty0 1--18, 2020.

\bibitem[Chopite et~al.(2020)Chopite, Hullin, Wand, and Iseringhausen]{chopite2020deep}
Javier~Grau Chopite, Matthias~B Hullin, Michael Wand, and Julian Iseringhausen.
\newblock Deep non-line-of-sight reconstruction.
\newblock In \emph{Proceedings of the IEEE/CVF Conference on Computer Vision and Pattern Recognition}, pages 960--969, 2020.

\bibitem[Dao et~al.(2022)Dao, Fu, Ermon, Rudra, and R{\'e}]{dao2022flashattention}
Tri Dao, Dan Fu, Stefano Ermon, Atri Rudra, and Christopher R{\'e}.
\newblock Flashattention: Fast and memory-efficient exact attention with io-awareness.
\newblock \emph{Advances in Neural Information Processing Systems}, 35:\penalty0 16344--16359, 2022.

\bibitem[Faccio et~al.(2020)Faccio, Velten, and Wetzstein]{faccio2020non}
Daniele Faccio, Andreas Velten, and Gordon Wetzstein.
\newblock Non-line-of-sight imaging.
\newblock \emph{Nature Reviews Physics}, 2\penalty0 (6):\penalty0 318--327, 2020.

\bibitem[Geng et~al.(2021)Geng, Hu, Chen, et~al.]{geng2021recent}
Ruixu Geng, Yang Hu, Yan Chen, et~al.
\newblock Recent advances on non-line-of-sight imaging: Conventional physical models, deep learning, and new scenes.
\newblock \emph{APSIPA Transactions on Signal and Information Processing}, 11\penalty0 (1), 2021.

\bibitem[Gupta et~al.(2012)Gupta, Willwacher, Velten, Veeraraghavan, and Raskar]{gupta2012reconstruction}
Otkrist Gupta, Thomas Willwacher, Andreas Velten, Ashok Veeraraghavan, and Ramesh Raskar.
\newblock Reconstruction of hidden 3d shapes using diffuse reflections.
\newblock \emph{Optics express}, 20\penalty0 (17):\penalty0 19096--19108, 2012.

\bibitem[Heide et~al.(2019)Heide, O'Toole, Zang, Lindell, Diamond, and Wetzstein]{heide2019non}
Felix Heide, Matthew O'Toole, Kai Zang, David~B Lindell, Steven Diamond, and Gordon Wetzstein.
\newblock Non-line-of-sight imaging with partial occluders and surface normals.
\newblock \emph{ACM Transactions on Graphics}, 38\penalty0 (3):\penalty0 1--10, 2019.

\bibitem[Isogawa et~al.(2020)Isogawa, Chan, Yuan, Kitani, and O’Toole]{isogawa2020efficient}
Mariko Isogawa, Dorian Chan, Ye Yuan, Kris Kitani, and Matthew O’Toole.
\newblock Efficient non-line-of-sight imaging from transient sinograms.
\newblock In \emph{European conference on computer vision}, pages 193--208. Springer, 2020.

\bibitem[Kirmani et~al.(2009)Kirmani, Hutchison, Davis, and Raskar]{kirmani2009}
Ahmed Kirmani, Tyler Hutchison, James Davis, and Ramesh Raskar.
\newblock Looking around the corner using transient imaging.
\newblock In \emph{IEEE 12th International Conference on Computer Vision}, pages 159--166. IEEE, 2009.

\bibitem[Li et~al.(2023)Li, Zhang, Ye, Xu, and Xiong]{li2023deep}
Yue Li, Yueyi Zhang, Juntian Ye, Feihu Xu, and Zhiwei Xiong.
\newblock Deep non-line-of-sight imaging from under-scanning measurements.
\newblock In \emph{Advances in Neural Information Processing Systems}, pages 1--12, 2023.

\bibitem[Lindell et~al.(2019)Lindell, Wetzstein, and O'Toole]{lindell2019fk}
David~B Lindell, Gordon Wetzstein, and Matthew O'Toole.
\newblock Wave-based non-line-of-sight imaging using fast fk migration.
\newblock \emph{ACM Transactions on Graphics (ToG)}, 38\penalty0 (4):\penalty0 1--13, 2019.

\bibitem[Liu et~al.(2019)Liu, Guill{\'e}n, La~Manna, Nam, Reza, Huu~Le, Jarabo, Gutierrez, and Velten]{liu2019PF}
Xiaochun Liu, Ib{\'o}n Guill{\'e}n, Marco La~Manna, Ji~Hyun Nam, Syed~Azer Reza, Toan Huu~Le, Adrian Jarabo, Diego Gutierrez, and Andreas Velten.
\newblock Non-line-of-sight imaging using phasor-field virtual wave optics.
\newblock \emph{Nature}, 572\penalty0 (7771):\penalty0 620--623, 2019.

\bibitem[Liu et~al.(2020)Liu, Bauer, and Velten]{liu2020NC}
Xiaochun Liu, Sebastian Bauer, and Andreas Velten.
\newblock Phasor filed diffraction based reconstruction for fast non-line-of-sight imaging systems.
\newblock \emph{Nature communications}, 11:\penalty0 1645, 2020.

\bibitem[Liu et~al.(2023{\natexlab{a}})Liu, Wang, Xiao, Fu, Qiu, and Shi]{liu2023fewshot}
Xintong Liu, Jianyu Wang, Leping Xiao, Xing Fu, Lingyun Qiu, and Zuoqiang Shi.
\newblock Few-shot non-line-of-sight imaging with signal-surface collaborative regularization.
\newblock In \emph{Proceedings of the IEEE/CVF Conference on Computer Vision and Pattern Recognition}, pages 13303--13312, 2023{\natexlab{a}}.

\bibitem[Liu et~al.(2023{\natexlab{b}})Liu, Wang, Xiao, Shi, Fu, and Qiu]{liu2023NC}
Xintong Liu, Jianyu Wang, Leping Xiao, Zuoqiang Shi, Xing Fu, and Lingyun Qiu.
\newblock Non-line-of-sight imaging with arbitrary illumination and detection pattern.
\newblock \emph{Nature Communications}, 14\penalty0 (1):\penalty0 3230, 2023{\natexlab{b}}.

\bibitem[Liu et~al.(2022)Liu, Ning, Cao, Wei, Zhang, Lin, and Hu]{liu2022video}
Ze Liu, Jia Ning, Yue Cao, Yixuan Wei, Zheng Zhang, Stephen Lin, and Han Hu.
\newblock Video swin transformer.
\newblock In \emph{Proceedings of the IEEE/CVF conference on computer vision and pattern recognition}, pages 3202--3211, 2022.

\bibitem[Metzler et~al.(2021)Metzler, Lindell, and Gordon]{metzler2021keyhole}
Christopher~A. Metzler, David~B. Lindell, and Wetzstein Gordon.
\newblock Keyhole imaging: Non-line-of-sight imaging and tracking of moving objects along a single optical path.
\newblock \emph{IEEE Transactions on Computational Imaging}, 7:\penalty0 1--12, 2021.

\bibitem[Mu et~al.(2022)Mu, Mo, Peng, Liu, Nam, Raghavan, Velten, and Li]{mu2022NLOS3d}
Fangzhou Mu, Sicheng Mo, Jiayong Peng, Xiaochun Liu, Ji~Hyun Nam, Siddeshwar Raghavan, Andreas Velten, and Yin Li.
\newblock Physics to the rescue: Deep non-line-of-sight reconstruction for high-speed imaging.
\newblock \emph{IEEE Transactions on Pattern Analysis and Machine Intelligence}, 2022.

\bibitem[Nam et~al.(2021)Nam, Brandt, Bauer, Liu, Renna, Tosi, Sifakis, and Velten]{nam2021NC}
Ji~Hyun Nam, Eric Brandt, Sebastian Bauer, Xiaochun Liu, Marco Renna, Alberto Tosi, Eftychios Sifakis, and Andreas Velten.
\newblock Low-latency time-of-flight non-line-of-sight imaging at 5 frames per second.
\newblock \emph{Nature communications}, 12:\penalty0 6526, 2021.

\bibitem[O’Toole et~al.(2018)O’Toole, Lindell, and Wetzstein]{otoole2018LCT}
Matthew O’Toole, David~B Lindell, and Gordon Wetzstein.
\newblock Confocal non-line-of-sight imaging based on the light-cone transform.
\newblock \emph{Nature}, 555\penalty0 (7696):\penalty0 338--341, 2018.

\bibitem[Pan et~al.(2022)Pan, Li, Gao, Wang, Shen, Liu, Wu, Yu, and Li]{pan2022onsite}
Zhengqing Pan, Ruiqian Li, Tian Gao, Zi Wang, Siyuan Shen, Ping Liu, Tao Wu, Jingyi Yu, and Shiying Li.
\newblock Onsite non-line-of-sight imaging via online calibration.
\newblock \emph{IEEE Photonics Journal}, 14\penalty0 (5):\penalty0 1--11, 2022.

\bibitem[Pei et~al.(2021)Pei, Zhang, Deng, Xu, Wu, David, Li, Qiao, Fang, and Dai]{pei2021dynamic}
Chengquan Pei, Anke Zhang, Yue Deng, Feihu Xu, Jiamin Wu, U David, Lei Li, Hui Qiao, Lu Fang, and Qionghai Dai.
\newblock Dynamic non-line-of-sight imaging system based on the optimization of point spread functions.
\newblock \emph{Optics Express}, 29\penalty0 (20):\penalty0 32349--32364, 2021.

\bibitem[Shen et~al.(2021)Shen, Wang, Liu, Pan, Li, Gao, Li, and Yu]{shen2021NeTF}
Siyuan Shen, Zi Wang, Ping Liu, Zhengqing Pan, Ruiqian Li, Tian Gao, Shiying Li, and Jingyi Yu.
\newblock Non-line-of-sight imaging via neural transient fields.
\newblock \emph{IEEE Transactions on Pattern Analysis and Machine Intelligence}, 43\penalty0 (7):\penalty0 2257--2268, 2021.

\bibitem[Tsai et~al.(2019)Tsai, Sankaranarayanan, and Gkioulekas]{tsai2019beyond}
Chia-Yin Tsai, Aswin~C Sankaranarayanan, and Ioannis Gkioulekas.
\newblock Beyond volumetric albedo--a surface optimization framework for non-line-of-sight imaging.
\newblock In \emph{Proceedings of the IEEE/CVF conference on computer vision and pattern recognition}, pages 1545--1555, 2019.

\bibitem[Velten et~al.(2012)Velten, Willwacher, Gupta, Veeraraghavan, Bawendi, and Raskar]{velten2012NC}
Andreas Velten, Thomas Willwacher, Otkrist Gupta, Ashok Veeraraghavan, Moungi~G Bawendi, and Ramesh Raskar.
\newblock Recovering three-dimensional shape around a corner using ultrafast time-of-flight imaging.
\newblock \emph{Nature communications}, 3\penalty0 (1):\penalty0 745, 2012.

\bibitem[Wang et~al.(2023)Wang, Liu, Xiao, Shi, Qiu, and Fu]{wang2023non}
Jianyu Wang, Xintong Liu, Leping Xiao, Zuoqiang Shi, Lingyun Qiu, and Xing Fu.
\newblock Non-line-of-sight imaging with signal superresolution network.
\newblock In \emph{Proceedings of the IEEE/CVF Conference on Computer Vision and Pattern Recognition}, pages 17420--17429, 2023.

\bibitem[Wang et~al.(2022)Wang, Cao, Zhong, and Yuan]{wang2022spatial}
Lishun Wang, Miao Cao, Yong Zhong, and Xin Yuan.
\newblock Spatial-temporal transformer for video snapshot compressive imaging.
\newblock \emph{IEEE Transactions on Pattern Analysis and Machine Intelligence}, 45\penalty0 (7):\penalty0 9072--9089, 2022.

\bibitem[Xiaohua and Liang(2021{\natexlab{a}})]{feng2021toward}
Feng Xiaohua and Gao Liang.
\newblock Toward non-line-of-sight videography.
\newblock \emph{Optics and Photonics News}, 32\penalty0 (11):\penalty0 22--29, 2021{\natexlab{a}}.

\bibitem[Xiaohua and Liang(2021{\natexlab{b}})]{feng2021ultrafast}
Feng Xiaohua and Gao Liang.
\newblock Ultrafast light field tomography for snapshot transient and non-line-of-sight imaging.
\newblock \emph{Nature Communications}, 12:\penalty0 2179, 2021{\natexlab{b}}.

\bibitem[Xin et~al.(2019)Xin, Nousias, Kutulakos, Sankaranarayanan, Narasimhan, and Gkioulekas]{xin2019theory}
Shumian Xin, Sotiris Nousias, Kiriakos~N Kutulakos, Aswin~C Sankaranarayanan, Srinivasa~G Narasimhan, and Ioannis Gkioulekas.
\newblock A theory of fermat paths for non-line-of-sight shape reconstruction.
\newblock In \emph{Proceedings of the IEEE/CVF conference on computer vision and pattern recognition}, pages 6800--6809, 2019.

\bibitem[Ye et~al.(2024{\natexlab{a}})Ye, Hong, Su, Yuan, and Xu]{ye2024plugandplay}
Juntian Ye, Yu Hong, Xiongfei Su, Xin Yuan, and Feihu Xu.
\newblock Plug-and-play algorithms for dynamic non-line-of-sight imaging.
\newblock \emph{ACM Transactions on Graphics}, 43\penalty0 (5):\penalty0 1--12, 2024{\natexlab{a}}.

\bibitem[Ye et~al.(2021)Ye, Huang, Li, and Xu]{ye2021compressed}
Jun-Tian Ye, Xin Huang, Zheng-Ping Li, and Feihu Xu.
\newblock Compressed sensing for active non-line-of-sight imaging.
\newblock \emph{Optics Express}, 29\penalty0 (2):\penalty0 1749--1763, 2021.

\bibitem[Ye et~al.(2024{\natexlab{b}})Ye, Sun, Li, Zeng, Hong, Li, Huang, Xue, Yuan, Xu, Dou, and Pan]{Ye2024}
Jun-Tian Ye, Yi Sun, Wenwen Li, Jian-Wei Zeng, Yu Hong, Zheng-Ping Li, Xin Huang, Xianghui Xue, Xin Yuan, Feihu Xu, Xiankang Dou, and Jian-Wei Pan.
\newblock Real-time non-line-of-sight computational imaging using spectrum filtering and motion compensation.
\newblock \emph{Nature Computational Science}, 4:\penalty0 920–927, 2024{\natexlab{b}}.

\bibitem[Young et~al.(2024)Young, Batagoda, Zhang, Dave, Pediredla, Negrut, and Raskar]{young2024enhancing}
Aaron Young, Nevindu~M Batagoda, Harry Zhang, Akshat Dave, Adithya Pediredla, Dan Negrut, and Ramesh Raskar.
\newblock Enhancing autonomous navigation by imaging hidden objects using single-photon lidar.
\newblock \emph{arXiv preprint arXiv:2410.03555}, 2024.

\bibitem[Young et~al.(2020)Young, Lindell, Girod, Taubman, and Wetzstein]{young2020dlct}
Sean~I Young, David~B Lindell, Bernd Girod, David Taubman, and Gordon Wetzstein.
\newblock Non-line-of-sight surface reconstruction using the directional light-cone transform.
\newblock In \emph{Proceedings of the IEEE/CVF Conference on Computer Vision and Pattern Recognition}, pages 1407--1416, 2020.

\bibitem[Yu et~al.(2023)Yu, Shen, Wang, Huang, Wang, Peng, Xia, Liu, Li, and Li]{yu2023enhancing}
Yanhua Yu, Siyuan Shen, Zi Wang, Binbin Huang, Yuehan Wang, Xingyue Peng, Suan Xia, Ping Liu, Ruiqian Li, and Shiying Li.
\newblock Enhancing non-line-of-sight imaging via learnable inverse kernel and attention mechanisms.
\newblock In \emph{IEEE/CVF International Conference on Computer Vision}, pages 10563--10573, 2023.

\bibitem[Zhang et~al.(2024)Zhang, Guo, Zhu, Huang, Chen, Liu, Bai, Lam, and Han]{Zhang2024}
Wenjun Zhang, Enlai Guo, Shuo Zhu, Chenyang Huang, Lijia Chen, Lingfeng Liu, Lianfa Bai, Edmund~Y. Lam, and Jing Han.
\newblock Real-time scan-free non-line-of-sight imaging.
\newblock \emph{APL Photonics}, 9\penalty0 (12), 2024.

\end{thebibliography}
}

\end{document}